%% file: arxiv.tex
\documentclass[runningheads]{llncs}
\usepackage{fancyhdr}
\usepackage{graphicx}
\usepackage[misc,geometry]{ifsym} 
\usepackage{amsmath}
\usepackage{amssymb}
\begin{document}
\title{Efficient Human Pose Estimation with Depthwise Separable Convolution and Person Centroid Guided Joint Grouping}
%\footnote{This paper has been accepted by 3rd Chinese Conference on Pattern Recognition and Computer Vision, PRCV 2020.}
%
\titlerunning{Efficient Human Pose Estimation}
% If the paper title is too long for the running head, you can set
% an abbreviated paper title here
%
\author{Jie Ou \and
Hong Wu}
\authorrunning{J.~Ou and H.~Wu}
% First names are abbreviated in the running head.
% If there are more than two authors, 'et al.' is used.
%
\institute{School of Computer Science and Engineering,\\
University of Electronic Science and Technology of China,\\
Chengdu 611731, China\\
\email{oujieww6@gmail.com, hwu@uestc.edu.cn}\\
}
\maketitle              % typeset the header of the contribution
\thispagestyle{fancy}            %更改plain状态，首页格式设为fancy
\fancyhead{}                     %清除以前的命令
%\rhead{righthead}                %页眉右侧内容
\cfoot{This paper has been accepted by 3rd Chinese Conference on Pattern Recognition and Computer Vision, PRCV 2020.}                 %页脚左侧内容
%\cfoot{\quad}                    %没找到清除页码的，直接用空格覆盖住页脚中间页码
 
\renewcommand{\headrulewidth}{0pt}      %把页眉线的宽度设为零，即去掉页眉线
\renewcommand{\footrulewidth}{0.5pt}
 
\pagestyle{empty} 
\input{section/abstract}
\input{section/introduction}
\input{section/relatex}
\input{section/method}
\input{section/exp}
\input{section/conc}
%
%
%
\section{Acknowledgement}
This work was supported in part by the Sichuan Science and Technology Program, China, under grants No.2020YFS0057, and the Fundamental Research Funds for the Central Universities under Project ZYGX2019Z015.
%
% ---- Bibliography ----
%
% BibTeX users should specify bibliography style 'splncs04'.
% References will then be sorted and formatted in the correct style.
%
\bibliographystyle{splncs04}
\bibliography{mybib2}

\end{document}

%% file: section/abstract.tex
\begin{abstract}
In this paper, we propose efficient and effective methods for 2D human pose estimation. A new ResBlock is proposed based on depthwise separable convolution and is utilized instead of the original one in Hourglass network. It can be further enhanced by replacing the vanilla depthwise convolution with a mixed depthwise convolution. Based on it, we propose a bottom-up multi-person pose estimation method. A rooted tree is used to represent human pose by introducing person centroid as the root which connects to all body joints directly or hierarchically. Two branches of sub-networks are used to predict the centroids, body joints and their offsets to their parent nodes. Joints are grouped by tracing along their offsets to the closest centroids. Experimental results on the MPII human dataset and the LSP dataset show that both our single-person and multi-person pose estimation methods can achieve competitive accuracies with low computational costs.

\keywords{Human Pose Estimation \and  Depthwise separable convolution  \and Hourglass network \and Joint Grouping}
\end{abstract}

%% file: section/introduction.tex
\section{Introduction}
Human pose estimation aims to locate human body joints from a single monocular image. It is a challenge and fundamental task  in many visual applications, e.g. surveillance, autonomous driving, human-computer interaction, etc.  In the last few years, considerable progress on human pose estimation has been achieved by deep learning based approaches~\cite{newell2016stacked,xiao2018simple,sun2019deep,nie2018pose,nie2019single,zhang2019fast}.
% Human pose estimation is the task of detecting the joints of a person in an image or video frame. It is a key to many visual applications, \emph{e.g.} surveillance, autonomous driving, human-computer interaction, \emph{etc}. It is very challenging to recognize human pose in the wild, due to occlusions, high flexibility of body joints, challenging view condition, various clothing, \emph{etc}. In recent years, considerable progress on human pose estimation has been achieved by deep learning based approaches~\cite{newell2016stacked,xiao2018simple,sun2019deep,nie2018pose,nie2019single,zhang2019fast}.

Most existing research works on human pose estimation focus on improving the accuracy and develop deep networks with large model size and low computational efficiency, which prohibits their practical application. To adopt deep networks in real-time applications and/or on limited resource devices, the model should be compact and computational efficient. Inception module~\cite{szegedy2016rethinking} is used to build deeper networks without increase model size and computational cost. 
Depthwise separable convolution~\cite{chollet2017xception,howard2017mobilenets,tan2019mixconv}, has been utilized as the key building block in many successful efficient CNNs.
In this paper, we follow these successful design principles to develop efficient deep networks for human pose estimation.
%Recently, many model compression methods have been proposed, and they can be mainly classified into two categories: compressing existing deep networks and designing new efficient architectures. The compressing approach usually bases on traditional compression techniques such as hashing~\cite{chen2015compressing}, Huffman coding~\cite{han2015deep}, factorization~\cite{jaderberg2014speeding}, pruning~\cite{see2016compression}, and product quantization~\cite{wu2016quantized}. The second approach actually has a longer history. Inception module~\cite{ioffe2015batch,szegedy2016rethinking} is used to build deeper networks without increase model size and computational cost. Depthwise separable convolution has been utilized as the key building block in many successful efficient CNNs, such as Xception module~\cite{chollet2017xception}, MobileNets~\cite{howard2017mobilenets,sandler2018mobilenetv2} and ShuffleNets~\cite{zhang2018shufflenet,ma2018shufflenet}. MixConv~\cite{tan2019mixconv} further extends vanilla depthwise convolution by using different kernel sizes to improve its representation capability. Besides these two approaches, there are some works trying to reduce the inference time, \emph{e.g.} knowledge distillation. The knowledge distillation methods~\cite{bucilu2006model,hinton2015distilling,radosavovic2018data} train small student networks to reproduce the output of large teacher networks to reduce inference-time
%costs. In this paper, we follow some successful design principles to develop efficient deep networks for human pose estimation.

For multi-person pose estimation, it is needed to distinguishing poses of different persons. The approaches can mainly be divided into two categories: top-down strategy and bottom-up strategy. The top-down approaches~\cite{papandreou2017towards,iqbal2016multi,fang2017rmpe,xiao2018simple,sun2019deep} employ detectors to localize person instances and then apply joint detector to each person instance. Each step of top-down approaches requires a very large amount of calculations, and the run-time of the second step is proportional to the number of person. In contrast, the bottom-up approaches~\cite{papandreou2018personlab,pishchulin2016deepcut,levinkov2017joint,insafutdinov2016deepercut,insafutdinov2017arttrack,cao2017realtime} detect all the body joints for only once and then group/allocate them into different persons. However, they suffer from very high complexity of joint grouping step, which usually involves solving a NP-hard graph partition problem. Different methods have been proposed to reduce the grouping time. Recently, some one-stage multi-person pose estimation approaches~\cite{nie2019single,sekii2018pose,tian2019directpose} have been proposed, but their performance lag behind the two-stage ones. In this paper, we also focus on the bottom-up strategy. 

In this paper, we propose efficient and effective methods for 2D human pose estimation. A new ResBlock is proposed with two depthwise separable convolutions and a squeeze-and-excitation (SE) module and utilized in place of the original ResBlock in Hourglass network. Its representation capability is further enhanced by replacing the vanilla depthwise convolution with a mixed depthwise convolution. The new Hourglass networks is very light-weighted and can be directly applied to single-person pose estimation. 
Base on this backbone network, we further propose a new bottom-up multi-person pose estimation method. A rooted tree is used to represent human pose by introducing person centroid as the root which connecting to all the joints directly or hierarchically. Two branches of sub-networks are used to predict the centroids, body joints and their offsets to their parent nodes. Joints are grouped by tracing along their offsets to the closest centroids. Our single-person pose estimation method is evaluated on MPII Human Pose dataset~\cite{andriluka20142d} and Leeds Sports Pose dataset~\cite{johnson2010clustered}. It achieves competitive accuracy with only 4.7 GFLOPs. 
Our multi-person pose estimation method is evaluated on MPII Human Pose Multi-Person dataset~\cite{andriluka20142d}, and achieves competitive accuracy with only 13.6 GFLOPs. %The main contributions of this paper are summarized as follows:  
% \begin{enumerate}
% \item We propose a new ResBlock which is composed of two depthwise separable convolutions and a squeeze-and-excitation module, and use it to build a lightweight Hourglass network.
% \item We further enhance the proposed ResBlock by replacing the vanilla depthwise convolution with a mixed depthwise convolution.
% \item Based on the proposed light-weight Hourglass network, we develop a efficient bottom-up multi-person pose estimation method.
% \item Both our single-person pose estimation method and multi-person pose estimation method can achieve competitive accuracies with low computational costs.
% \end{enumerate}

%The rest of the paper is organized as follows. Under Section II, we introduce some works related to our research. Then, the proposed light-weight Hourglass network is introduced in section III. Single-person estimation and Multi-person estimation methods are given in Section IV and V. Section VI shows the experimental results and analysis. Finally, we conclude this paper in Section VII.

%% file: section/relatex.tex
\section{Related Works}

\subsection{Efficient Neural Networks}
% To adopt deep neural networks in real-time applications and/or on resource-constrained devices, many research works have been devoted to build efficient neural networks with acceptable performance. Depthwise separable convolution was originally presented in~\cite{sifre2014rigid}. It can achieve a good balance between the representation capability and computational efficiency, and  has been utilized as the key building block in many successful efficient CNNs, such as Xception~\cite{chollet2017xception} and  MobileNets~\cite{sandler2018mobilenetv2}. MixConv~\cite{tan2019mixconv} extends vanilla depthwise convolution by partitioning channels into multiple groups and apply different kernel sizes to each of them, and achieves better representation capability. 
To adopt deep neural networks in real-time applications and/or on resource-constrained devices, many research works have been devoted to build efficient neural networks with acceptable performance. Depthwise separable convolution was originally presented in~\cite{sifre2014rigid}. It can achieve a good balance between the representation capability and computational efficiency, and  has been utilized as the key building block in many successful efficient CNNs, such as Xception~\cite{chollet2017xception}, MobileNets~\cite{howard2017mobilenets,sandler2018mobilenetv2} and ENAS~\cite{pham2018efficient}. MixConv~\cite{tan2019mixconv} extends vanilla depthwise convolution by partitioning channels into multiple groups and apply different kernel sizes to each of them, and achieves better representation capability. 

\subsection{Multi-person pose estimation}
\textbf{Top-down methods}. Top-down multi-person pose estimation methods first detect people by a human detector (\emph{e.g.} Faster-RCNN\cite{ren2015faster} ), then run a single-person pose estimator on the cropped image of each person to get the final pose predictions. Representative top-down methods include PoseNet~\cite{papandreou2017towards}, RMPE~\cite{fang2017rmpe}, Mask R-CNN~\cite{he2017mask}, CPN~\cite{chen2018cascaded} and MSRA~\cite{xiao2018simple}. However, top-down methods depend heavily on the human detector, and their inference time will significantly increase if many people appear together.

%[35] G. Papandreou, T. Zhu, N. Kanazawa, A. Toshev, J. Tompson, C. Bregler, and K. Murphy. Towards accurate multi-person pose estimation in the wild. In CVPR, volume 3, page 6, 2017. 2, 3
%[10] H. Fang, S. Xie, and C. Lu. Rmpe: Regional multiperson pose estimation. 2017 IEEE International Conference on Computer Vision (ICCV), pages 2353–2362,
%2017. 2
%[18] K. He, G. Gkioxari, P. Doll´ar, and R. Girshick. Mask r-cnn. In Computer Vision (ICCV), 2017 IEEE International Conference on, pages 2980–2988. IEEE, 2017.
%[6] Y. Chen, Z. Wang, Y. Peng, Z. Zhang, G. Yu, and J. Sun. Cascaded pyramid network for multi-person pose estimation. In Proceedings of the IEEE Conference on Computer Vision and Pattern Recognition, pages 7103–7112, 2018.
%[44] B. Xiao, H. Wu, and Y. Wei. Simple baselines for human pose estimation and tracking. arXiv preprint arXiv:1804.06208, 2018.

\textbf{Bottom-up methods}. Bottom-up methods detect the human joints of all persons at once, and then allocate these joints to each person based on various joint grouping methods. However, they suffer from very high complexity of joint grouping step, which usually involves solving a NP-hard graph partition problem. DeepCut~\cite{pishchulin2016deepcut} and DeeperCut~\cite{insafutdinov2016deepercut} solve the joint grouping with an integer linear program which results in the order of hours to process a single image. Later works drastically reduce prediction time by using greedy decoders in combination with additional tools. Cao~\emph{et~al.}~\cite{cao2017realtime} proposed part affinity fields to encode location and orientation of limbs. Newell and Deng~\cite{newell2017associative} presented the associative embedding for grouping joint candidates. PPN~\cite{nie2018pose} performs dense regressions from global joint candidates within a embedding space of person centroids to generate person detection and joint grouping. But it need to adopt the Agglomerative Clustering algorithm~\cite{bourdev2009poselets} to determine the person centroids. In this paper, we avoid the time-consuming clustering by regressing the person centroids together with body joints and using them to guide the joint grouping. 

Recently, Nie~\emph{et~al.}~\cite{nie2019single} proposed a one-stage multi-person pose estimation method (SPM) which predicts root joints (person centroids) and joint displacements directly. Although both SPM and our method predict person centroid, they use centroid plus displacements to recover joints and we use centroid to guide joint grouping. We argue that joints can be predicted more precisely than its displacements.

%% file: section/method.tex
\section{The proposed light-weight Hourglass network}
\subsection{Hourglass network} 
\begin{figure}[!t] % picture
    \centering
    \includegraphics[width=12cm]{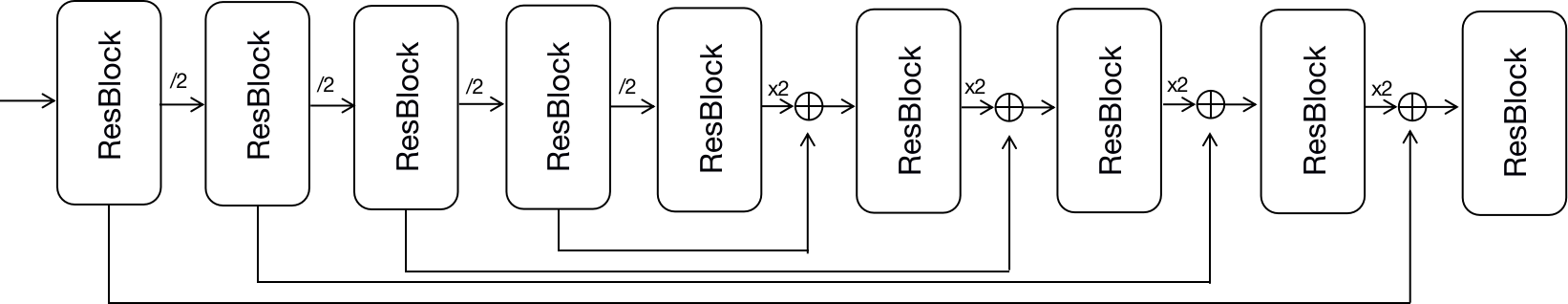}
    \caption{An illustration of a Hourglass module, ``/2'' means downsampling operation and ``x2'' means upsampling operation.
}\label{fig:1}
\end{figure}

\begin{figure}[!t] % picture
    \centering
    \includegraphics[width=12cm]{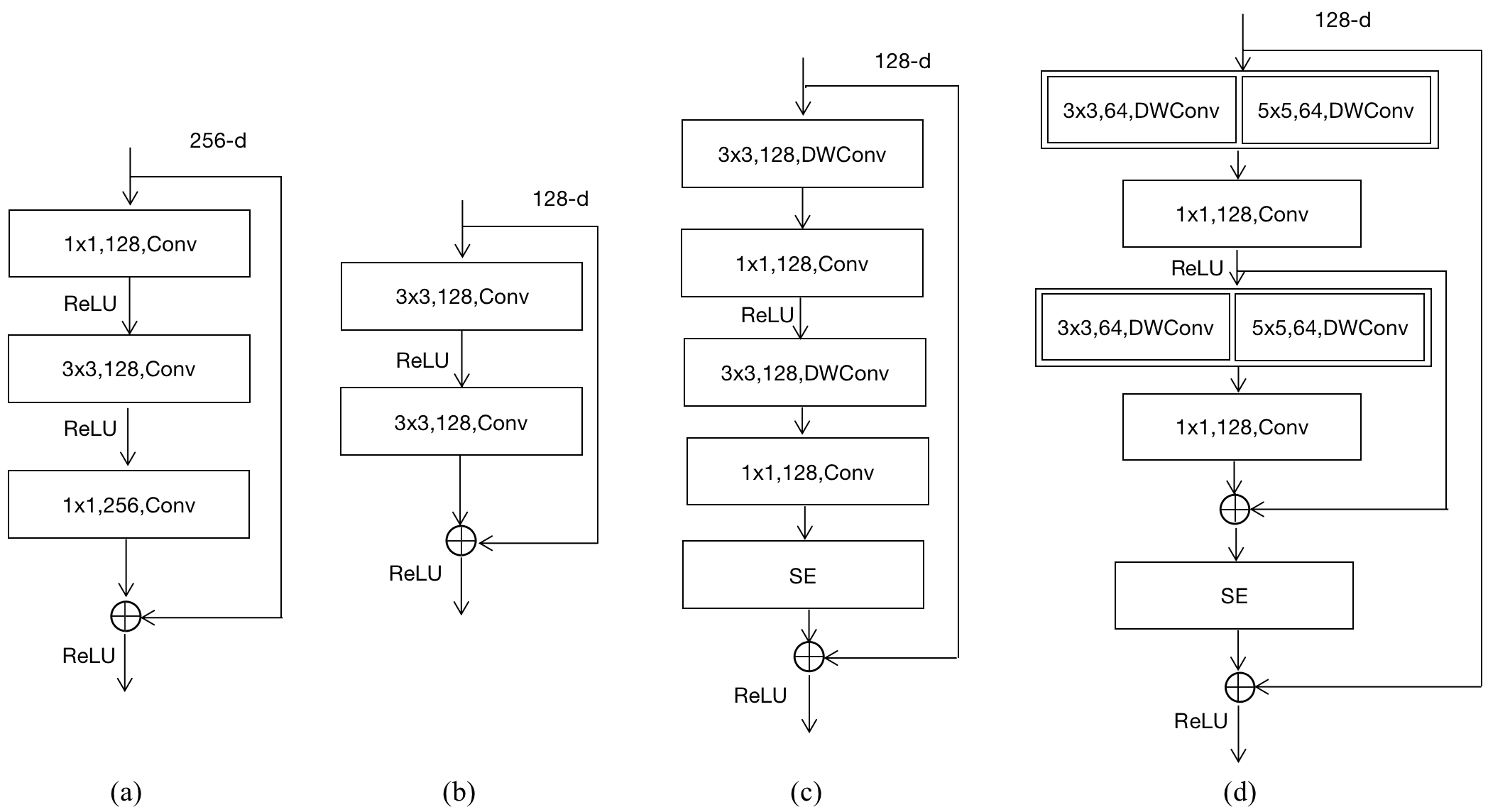}
    \caption{Architecture of the ResBlocks. (a) The original ResBlock in Hourglass\cite{newell2016stacked}. (b) another kind of ResBlock proposed in ResNet\cite{he2016deep}. (c) DS-ResBlock corresponding to (b), (d) replace the DWConv in (c) with Mix-Cov.
}\label{fig:2}
\end{figure}

Although Hourglass network  has been utilized in many human pose estimation methods \cite{zhang2019fast,nie2018human,nie2018pose,nie2019single}, it is hard to been adapted in practical applications due to its large model size. Original Hourglass network consists of eight stacked hourglass modules, whose structure is illustrated in Fig.~\ref{fig:1}. %A hourglass module is composed of a series of Residual Blocks (ResBlocks) with downsampling operations followed by upsampling operations. 
The ResBlock used in original Hourglass network has a bottleneck structure (Fig.~\ref{fig:2}(a)). In this paper, we try to improve the efficiency of Hourglass network by replacing the original ResBlocks with the light-weight ones (Fig.~\ref{fig:1}). %by applying the depthwise separabe convolution and the mixed depthwise convolution. 
The proposed Hourglass network is called DS-Hourglass network. More details are given as follows.

%where a $1\times1$ convolution is first used to reduce the channel size from 256 to 128, then a $3\times3$ convolution is applied, finally, a $1\times1$ convolution is used to restore the channel size from 128 to 256. In this paper, we try to improve the efficiency of Hourglass network by replacing the original ResBlock with light-weight ones (Fig.~\ref{fig:1}). The proposed light-weight ResBlocks are called DS-ResBlocks, which are designed by applying the depthwise separabe convolution and the mixed depthwise convolution. And the derived Hourglass network is called DS-Hourglass network. More details are given as follows.

%Each box in the figure is a residual module as seen in Fig.~\ref{fig:2}. 
\subsection{Depthwise Separable Convolutions} 
A depthwise separable convolution decomposes a standard convolutional operation into a depthwise convolution (capture the spatial correlation) followed by a pointwise convolution (capture the cross-channel correlation). 

A standard convolution operation needs $ c_{1} \times c_{2} \times k \times k$ parameters and about $ h\times w \times c_{1} \times c_{2} \times  k \times k$ computational cost, where $h\times w$, $c_{1}/c_{2}$ and $k\times k$ are the spatial size of input and output feature maps, the number of input/output feature channels and the convolutional kernel size, respectively. While a depthwise separable convolution operation only needs $ c_{1} \times k \times k + c_{1}\times c_{2}$ parameters and about $ h\times w \times c_{1} (k^2 + c_{2})$ computational cost. For example, if we set $k$ to 3 and $c_{2}$ to 128, the number of parameters and the computational cost of the depthwise separable convolution is only about $1/9$ of the corresponding standard convolution. %This is the case in Fig.~\ref{fig:2}(c).

\subsection{Light-weight ResBlock} 
To develop efficient ResBlock, we first reduce the number of its input/output feature channels from 256 to 128, and use two stacked $3\times 3$ convolutions (Fig.~\ref{fig:2}(b)). Then, we replace the two standard $3\times 3$ convolutions with two depthwise separable convolutions followed with a squeeze-and-excitation (SE) block to get a light-weight ResBlock (Fig.~\ref{fig:2}(c)). The SE block is very efficicent and used to relocate features and strengthen features. To capture the information of different scales, we further replace the depthwise convolution with a mixed depthwise convolution (MixConv~\cite{tan2019mixconv}) to get another version of light-weight ResBlock (Fig.~\ref{fig:2}(d)). In MixConv, the input feature channels are first split into groups, then depthwise convolutions with different kernel sizes are applied to different groups, finally, the output of each depthwise convolution are concatenated. In this paper, we apply kernels of $3\times 3$ and $5\times 5$ to two groups of channels respectively to trade-off the representation capability and the computational costs. In our study, we found that adding a skip connection around the second depthwise separable convolution can make the training of this ResBlock (Fig.~\ref{fig:2}(d)) more stable.

\section{Multi-Person Pose Estimation}
\begin{figure}[!t] % picture
    \centering
    \includegraphics[width=12cm]{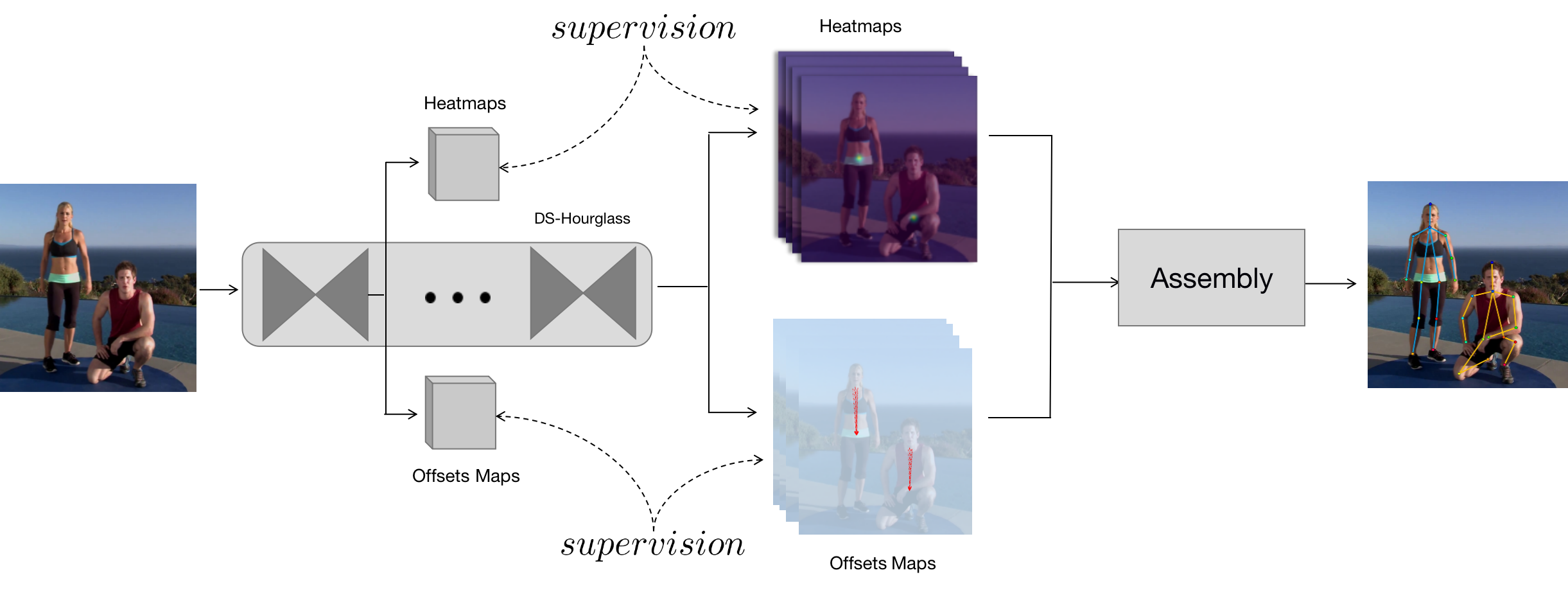}
    \caption{Our pipeline. Our method takes color image as input, the DS-Hourglass build with 8 stage and predict 2 sets, heatmaps and offsetmaps, finally return joints for each person in image at once by our efficient joint grouping algorithm.}\label{fig:3}
\end{figure}
Fig.~\ref{fig:3} illustrates the overall pipeline of our network. Our multi-person pose estimation method first predicts joints of all person at once, then the joint candidates are grouped into different persons. %To improve the efficiency of joint grouping, we introduce person centroid to guide the grouping. A rooted tree is used to represent human pose by using person centroid as the root which connecting to all the joints directly or hierarchically via offset prediction. 
To improve the efficiency of joint grouping, a rooted tree is used to represent human pose by using person centroid as the root which
connecting to all the joints directly or hierarchically. Another network branch is used to predict the offset from each joint to its parent node.
The person centroid is treated as a pseudo joint and predicted together with body joints.  After that, body joints are grouped by tracing along their offsets to the closed centroids. 
\begin{figure}[!t] % picture
    \centering
    \includegraphics[width=6.5cm,height=4.5cm]{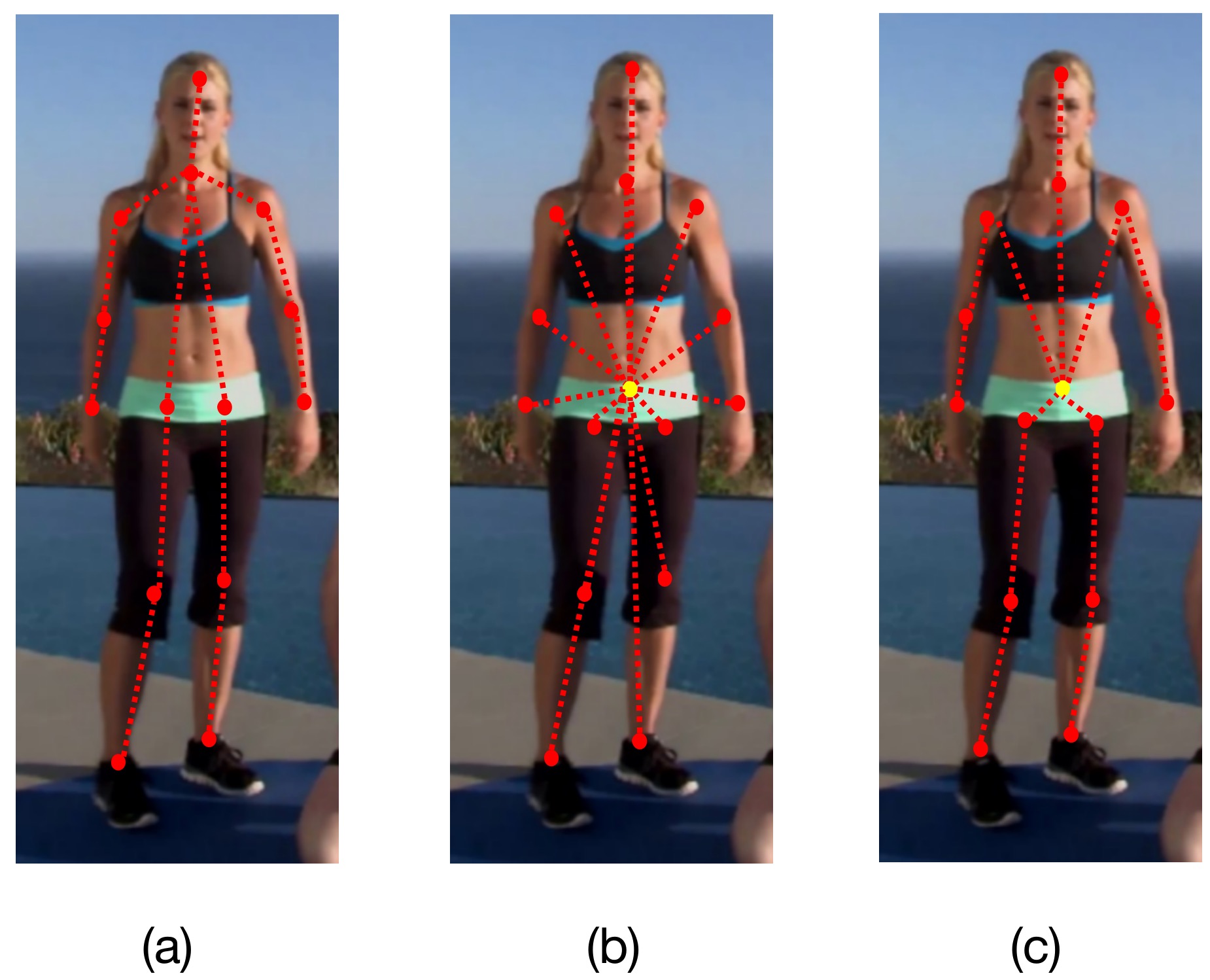}
    \caption{Pose representations: (a) kinematic structure; (b) centroid-rooted tree; (c) heirarchical centroid-rooted tree}\label{fig:4}
\end{figure}

\subsection{Centroid-rooted tree structure to define the offsets} 
%A rooted tree is used to represent human pose by introducing person centroid as the root which connecting to all the joints directly or hierarchically. 
%A rooted tree is used to represent human pose by introducing person centroid as the root which connecting to all the joints directly or hierarchically. This idea is similar to that in SPM~\cite{nie2019single}, the difference is that SPM use centroid plus displacements to recover joints and we use centroid to grouping the predicted joints. We argue that joints can be predicted more precisely than its displacements.
The centroid-rooted tree structure is illustrated in Fig.~\ref{fig:4}(b), where the person centriod (root node) directly connect to all body joints (leaf nodes). %And the offsets need to be predicted are defined as those from body joints to their parent node (the person centroid).
%This is our basic method,all of the parts of a person are related to the center of this person.That means  all of the $ V_{c}^{*}(x_{k}^{j},y_{k}^{j})$  point to the center of $ Person_{j} $ ,donated as $ Center_{j} $. Thus,in this way,we need extra joint detection confidence map for all $ Center $, $ P=\left (P_{1},P_{2},...,P_{K},P_{center}  \right ) $ and the  $ L=\left \{ \left \{ center,part_{i} \right \} $ $| i \in \left \{1,2,...,K \right\} $$\}$, as show in Fig.~\ref{fig:4}(b). It is  a very easy and simple way to encode the relation of part-to-part,moreover, because of its simple design, it is also very simple in assembly.
The drawback of this representation is that it leads to some long-range offsets which are hard to be precisely predicted, e.g. from ankle, knee and wrist to the centroid. %Due to the limited size of receptive field, these long-range offsets are hard to predict for CNN. 
To alleviate this problem, we further proposed a hierarchical centroid-rooted tree (Fig.~\ref{fig:4}(c)) based on the kinematic structure (Fig.~\ref{fig:4}(a)), where the long-range offsets are decomposed into short-range or middle-range offsets.
%To avoid long-range offset prediction, we further adopt a hierarchical centroid-rooted tree (Fig.~\ref{fig:4}(c)) to define the offsets, where the long-range offsets are decomposed into short-range or middle-range offsets, and have a corresponding body part/limb, based on the kinematic structure (Fig.~\ref{fig:4}(a)). 
%The drawback the above representation is that there are some long-range offsets need to be predicted, e.g. from ankle, knee and wrist to the centroid. Due to the limited size of receptive field, these long-range offsets are hard to predict for CNN. To avoid long-range offset prediction, we further adopt a hierarchical centroid-rooted tree (Fig.~\ref{fig:4}(c)) to define the offsets, where the long-range offsets are decomposed into short-range offsets based on the kinematic structure (Fig.~\ref{fig:4}(a)). The edges of the hierarchical centroid-rooted tree include $\{ (centroid,~neck),~(neck,~head),~(centroid,~right~shoulder),~(right~shoulder,$ $~right~elbow),~(right~elbow,~right~wirst),~(centroid,~right~hip),~(right~hip,$ $~right~knee),~(right~knee,~right~ankle),~(centroid,~left~shoulder)$ $,~(left~$ $shoulder,~left~elbow),~(left~elbow,~left~wirst),~(centroid,~left~hip),~(left~hip$ $,~left~knee),~(left~knee,~left~ankle) \}$. The offsets need to be predicted are defined as those from body joints to their parent node, and all the offsets are short-range or middle-range and have a corresponding body part/limb. This makes it easier for CNN to predict these offsets.

\subsection{Joint and Offset Prediction} 
Our network has two branches of sub-networks for joint and offset prediction. Both sub-networks have only one $1\times 1$ convolution and share the same feature maps  from  hourglass module. 

\textbf{Joint Prediction}. Our groud-truth heatmap is generated according to following equation,
%For joint prediction, there are two differences between our multi-person pose estimation method and our single-person pose estimation method. One is that person centroid is introduced and need to be predicted together with body joints. Another is that there are multiple persons in an input image, and the ground-truth heatmap should be generated in a different way. Our groud-truth heatmap is generated according to following equation,
\begin{equation}
H_{j}(x,y)=min(\sum_{i=1}^{N}exp(-\frac{((x,y)-(x_{j}^{i},y_{j}^{i}))^{2}}{2\sigma ^{2}}),1),
\end{equation}
where  $(x_{j}^{i},y_{j}^{i})$ is the coordinate of joint $j$ of person $i$, $N$ is the number of person in the image and $\sigma $ controls the spread of the peak, and minimum function is used to guarantee the value not greater than 1. 
%The MSE Loss in equation (8) is also used for joint prediction.

\textbf{Offset Prediction}.  we construct a dense offset map for each body joint as the ground-truth for offset prediction. We first construct a offset map $O_j^i$ for joint $j$ of person $i$ as: 
\begin{equation}
O_{j}^{i}(x, y)=\left\{\begin{array}{ll}
\frac{1}{Z}((x_{c}^{i}, y_{c}^{i})-(x, y)) & \text { if }(x, y) \in N_{j}^{i} \\
0 & \text { otherwise }
\end{array}\right.
\end{equation}
% \begin{equation}
% O_{j}^{i}(x, y)=\left\{\begin{array}{ll}
% \frac{1}{Z}((x_{c}^{i}, y_{c}^{i})-(x, y)), &  \text {if}(x, y) \in N_{j}^{i} \\
% 0 & , \text {otherwise}
% \end{array}\right.
% \end{equation}
%Following SPM~\cite{nie2019single}, we construct a dense offset map for each body joint as the ground-truth for offset prediction. The offset has different uses in SPM and our methods. In SPM, it is used to recover the coordinate of each body joint, but in our method, it is used to guide joint grouping. Therefore, offsets are point out from person centroids in SPM, but point toward person centroids directly or hierarchically in our method. We first construct a offset map $O_j^i$ for joint $j$ of person $i$ as 
% \begin{equation}
% O_j^i(x,y)=\begin{cases}
%  \frac{1}{Z}((x_{c}^{i},y_{c}^{i})-(x,y))& \text{ if } (x,y)\in N_{j}^{i} \\ 
%  0 & \text{ otherwise} 
% \end{cases},
% \end{equation}
where $ N_{j}^{i}=\left \{ (x,y)| \sqrt{(x,y)-(x_{j}^{i},y_{j}^{i}) }\leq \tau  \right \}$  denotes the area of neighbors of joint $j$ of person $i$, $(x_{c}^{i},y_{c}^{i})$ is the coordinate of the centroid of the person $i$, $ Z=\frac{1}{2}min(W,H) $ is a normalization coefficient, $ W $ and $ H $ are the width and height of the input image. If a location belongs to multiple people, these vectors are averaged. If the hierarchical centroid-rooted tree is used to represent human pose, we only need to replace the centroid in equation (2) with the parent node of the joint $j$ of the person $i$.
%Then, we average $O_j^i$ over all persons in the image to get the offset map $O_j$ for joint $j$.
% \begin{equation}
% O_j(x,y)=\frac{1}{M_j} \sum_{i=1}^{N} O_j^i(x,y), 
% \end{equation}
% where $M_j$ is the number of non-zero vectors at position $(x, y)$ over all persons. If the hierarchical centroid-rooted tree is used to represent human pose, we only need to replace the centroid in equation (2) with the parent node of joint $j$ of person $i$.

The MSE Loss is used for joint prediction, and Smooth L1-Loss is used for offset prediction. 

% The MSE Loss is used for joint prediction, and Smooth L1-Loss is used for offset prediction. The final loss is
% \begin{equation}
% Loss=\sum_{t=1}^{T}(Loss_{H}^{t}+Loss_{O}^{t}),
% \end{equation} where $Loss_{H}^{t}$ and $Loss_{O}^{t}$ are the losses of joint prediction and offset prediction at stage $t$ of Hourglass.

\subsection{Centroid-guided joint grouping} Based on the predicted person centroids, we develop a greedy method for joint grouping. First, we apply NMS to the heatmaps of the last stage of DS-Hourglass to get the coordinates of all candidate joints, and sort them in descending order of their score. 
%The candidates of each joint category are allocated one by one from high score to low score.
For the centroid-rooted tree representation, joint allocation is performed independently for each body joint. Given a candidate of joint $j$, its centroid's coordinate can be predicated as 
\begin{equation}
(\hat{x},\hat{y})_c = (x,y) + Z\times O_j(x,y),
\end{equation} where $(x,y)$ is the coordinate of the candidate joint, $(\hat{x},\hat{y})_c$ is the predicated coordinate of its person centroid, and $O_j$ is the offset map of joint $j$. %Then the predicated centroid is compared to each person centroid generated from heatmap. And this candidate is allocated to one of the person centroids based on the nearest-neighbour rule under the constrain that one person has only one candidate for one joint category.  
Then the predicated centroid is compared to each person centroid generated from heatmap and allocated to the nearest one under the constrain that one person has only one instance for each joint category.

For hierarchical centroid-rooted tree representation, the joints are also 
group-
ed hierarchically. Base on the intuition that the joints close to the torso can be predicted more reliably. We classify joints into three levels, the first level contains \textit{shoulder}, \textit{hip}, and \textit{neck}, the second level contains \textit{head}, \textit{elbow} and \textit{knee}, and third level contains \textit{wirst} and \textit{ankle}. The joint allocation is performed from the first level to the third level, each joint candidate is associated to its parent node in the tree structure according to equation (3) and the nearest-neighbour rule.

% \begin{figure}[!t] % picture
%     \centering
%     \includegraphics[width=8cm,height=5.5cm]{images/vis.png}
%     \caption{Example of Our Results}\label{fig:5}
% \end{figure}

%% file: section/exp.tex
\section{Experiments}
\begin{table}[!t]
 \caption{Comparison of ResBlocks on MPII single-person validation set}
  \centering
  
  \begin{tabular}{c||c|c|c|c}

  \hline
  Methods &Mean &Stages& Param &FLOPs\\

\hline
  \hline
  Hourglass~\cite{newell2016stacked}&90.52&8&26M&26.2G\\
  SBN~\cite{xiao2018simple}&89.6&-&68.6M&21G\\
  HRNet~\cite{sun2019deep}&90.3&-&28.5M&9.5G\\
  FPD~\cite{zhang2019fast}&89.04&4&3M&3.6G\\
  \hline
  DS-Hourglass*&88.71&8&2.9M&3.3G\\
  DS-Hourglass w/o SE&89.47&8&2.9M&4.7G\\
  DS-Hourglass&89.87&8&4.2M&4.7G\\
  DS-Hourglass(mix)&89.94&8&4.6M&4.8G\\
      \hline
  \end{tabular}
  \label{tab:table1}
\end{table}
\subsection{Experiment setup}
\textbf{Datasets} MPII Single-Person Dataset consists of around 25k images with annotations for multiple people providing 40k annotated samples (28k training, 11k testing) for single-person pose estimation. The MPII Multi-Person dataset consists of 3,844 and 1,758 groups of multiple interacting persons for training and testing. The LSP dataset has 11K training samples and 1K test samples, with 14 annotated joints for a person.
%respectively, we randomly selected 350 groups as validation set, only use 3494 groups for training.The LSP dataset has 11K training samples and 1K test samples, with 14 annotated joints for a person.

\textbf{Training Details} We randomly augmente the samples with rotation degrees in [-40, 40], scaling factors in [0.7, 1.3], translation offset in [-40, 40] and horizontally mirror, adopt $256\times 256$ as training size. %We implement our model with PyTorch~\cite{paszke2019pytorch}. 
The initial learning rate is 0.0025,  learning rate decay at step 150, 170, 200 and 230 with total 250 epochs by 0.5.

\subsection{Ablation Study}
\textbf{Ablation for ResBlocks} In Table 1, DS-Hourglass* uses  Fig.~\ref{fig:2}(c) with only one depthwise separable convolution. DS-Hourglass w/o SE uses  Fig.~\ref{fig:2}(c) without SE module. DS-Hourglass and DS-Hourglass(mix) use  Fig.~\ref{fig:2}(c) and Fig.~\ref{fig:2}(d) respectively. ResBlock with two depthwise separable convolution can improve 1.1\% in PCKh than ResBlock with only one, but with 1.4 GFLOPs computational cost increased. SE model can imporve 0.4\% in PCKh; Mixed depthwise convolution only bring very little improvement for single-person pose estimation. Compared with the excellent methods, our DS-Hourglass(mix) is 0.9\% higher than FPD. And 0.36\% lower than HRNet~\cite{sun2019deep}, However, the GFLOPs is only half of it and the parameters are only 16\% of HRNet.

\textbf{Ablation for Assembly methods} Table 2 indicated that our centroid-guided method improves 0.4\% over PPN, and our hierarchical centroid-guided method improves 1.8\% over PPN. DS-Hourglass(mix) improves 1\% over DS-Hourgalss with hierarchical centroid-guided assembly method, as it can handles multi-scale problem better, even better than PPN~\cite{nie2018pose} based on original Hourglass one (79.8\% vs 79.4\%), we can save 50GFLOPs and only need 21\% parameters of PPN.

\begin{table}[!t]
 \caption{Comparison of assembly methods on MPII multi-person validation set}
  \centering
  \begin{tabular}{c||c|c|c|c|c}
    
  \hline
  Model&Stage &Method&Mean & Param &FLOPs\\
      
    \hline
  \hline
  
  Hourglass &8&PPN~\cite{nie2018pose}&79.4&22M&62.9G\\
  \hline
  
  Hourglass&1&PPN&74.4&3.0M&10.8G\\
  Hourglass&1&Center&75.8&3.0M&10.9G\\
  Hourglass&1&Cent.Hier.&76.2&3.0M&10.9G\\
  \hline
  %DS-Hourglass(mix)&8&PPN~\cite{nie2018pose}&79.8&4.6M&13.6G\\
  %DS-Hourglass(mix)&8&Center&79.8&4.6M&13.6G\\
  DS-Hourglass&8&Cent.Hier.&78.8&4.4M&13.3G\\
  DS-Hourglass(mix)&8&Cent.Hier.&79.8&4.6M&13.6G\\

      \hline
  \end{tabular}

  \label{tab:table2}
\end{table}
\begin{table}[!t]

\begin{center}
\caption{Result on MPII Single-Person test set. $*$ means the use of extend dataset.}
\label{table:headings}
%\resizebox{11cm}{12mm}{
\begin{tabular}{c||c|c|c|c|c|c|c|c}
\hline
    Method & PCKh  & \textbf{Auxiliaries}& Stage&\textbf{Pre.} &Input size & Out. size & FLOPs & \#Param   \\
\hline
\hline

DeeperCut~\cite{insafutdinov2016deepercut}& 88.5  &-  &-&yes& $344\times 344$  & $ 43 \times 43 $  & 37G & 66M  \\
CPM*~\cite{Wei2016Convolutional}& 88.5   &- &6 &yes&$368 \times 368 $  & $46 \times 46 $  & 175G  & 31M   \\
SHG~\cite{newell2016stacked}&  90.9   &- & 8&no&$256 \times 256 $  & $ 64 \times 64 $  & 26.2G  & 26M  \\
PIL~\cite{nie2018human}&\textbf{92.4}  & Segment & 8&no&$ 256 \times 256 $ & $64 \times 64 $&  29.2G &26.4M\\
Sekii~\cite{sekii2018pose}&88.1&-&-&yes&$384 \times 384$ & $12 \times 12$ &6G&16M\\
FPD~\cite{zhang2019fast}&91.1  & 
Know. dist. &4 &yes&$ 256 \times 256 $&$ 64 \times 64 $ &\textbf{3.6G} &\textbf{3.2M}\\
HRNet~\cite{sun2019deep}&92.3&-&-&yes&$ 256 \times 256 $&$ 64 \times 64 $&9.5G&28.5M\\
\hline
\hline
\textbf{DS-Hourglass}& 91.5   &- & 8 &no&$ 256 \times 256 $&$ 64 \times 64 $  & 4.7G  & 4.2M \\

\hline
\end{tabular}

\end{center}
\end{table}

\subsection{Comparisons to State-Of-The-Art Methods}

\textbf{MPII Single-Person dataset} From Table 3, we can find that our method is very lightweight and efficient. Our model has greatly reduced the deployment cost, while still achieving a high PCKh of 91.5\%. Compared our method with the best performer, PIL~\cite{nie2018human}, the DS-Hourglass needs only 16\% of its computational cost but has only 0.9\% drop in PCKh. Our method outperforms FPD~\cite{zhang2019fast} (91.5\% vs 91.1\% AP) which needs  knowledge distillation and pretrained weights.  

%as Chu $et\quad al.$\cite{chu2017multi} which need 128GFLOPS and 58M paramters ,and our model only need 4.7GFLOPS  and 4.2M parameters .;Although our method does not achieve the state-of-the-art, there is not much loss in accuracy, less than 1\% from the best one
\textbf{LSP dataset} %In table 5, we compared our DS-Hourglass with a lot of existing methods with top reported performances on LSP test data. Our method also achieve the best 90.8\% PCK@0.2 accuracy on LSP dataset. We haven't listed the methods using additional datasets in their training process.
Our method also achieve the 90.8\% PCK@0.2 accuracy on LSP dataset which is same as FPD~\cite{zhang2019fast}, without using extra dataset. Because space is limited, the comparison is not listed in the form of a table.

\textbf{MPII Multi-Person dataset} In Table 4, we compare our method with the leading methods in recent years. It should be noted that our method does not use single-person pose estimation to refine the results. Research works~\cite{cao2017realtime,nie2018pose} have reported that the single-pose refinement can improve the result by about 2.6\%. However, the refinement is always time-consuming, so we did not use it. We get very competitive result 77.4\% achieve the state-of-the-art among the methods without refinement. Our model has only 4.6M parameters and needs 13.6 GFLOPs when using input size of $384\times 384$. To best of our knowledge, PPN~\cite{nie2018pose} is the state-of-the-art on MPII Multi-Person dataset, from our ablation Table 2, we can find that our method is better than PPN, and our method can reduced the computational cost by approximately 50GFLOPs.

Table 5 lists the results on MPII 288 test set, and our method gets 81.0\%, only 0.3\% lower than the best one~\cite{fieraru2018learning} which uses refinement. It can be found that our method has great advantages in the distal part of the body (\emph{e.g.} wrist, knee, ankle, \emph{etc}), as we use hierarchical centroid-rooted tree to avoid long-range offset predction.

%It is not fair to directly compare the runtimes. Just as the Python and Cuda version of OpenPose~\cite{cao2018openpose} have a large runtime difference, so we only give the network part  runtime and the Python-based assembly algorithm part runtime: 32ms/image and 3ms/person. If there is subsequent low-level optimization for depthwise convolution, our network will be more competitive.
%Based on the single-person pose estimation model, we improved the model as mentioned in 3.2, the residual module was changed to the  Figure 2(d).；
%The main purpose is that for the bottom-up multi-person pose estimation method, the model needs to process target people of different scales in the same image at once, so we hope to make the receptive field of the network more dynamic through this way. Inner skip connection allows the model to decide how to use the second convolution  during the learning process.We put some details in the ablation experimental part.network structure and assembly algorithm; In table3, $*$ means their result refined by  the single-person pose estimation .;this will also lose the advantage of the bottom-up approach. I;The method we proposed can reach almost the state-of-the-art, and can achieve real-time.
\begin{table}[!t]
 \caption{Results on the Full MPII Multi-Person test set.}
  \centering
  %\resizebox{10cm}{16.5mm}{
  \begin{tabular}{c||ccccccc||cc}
      \hline
  Methods &Head & Shoulder & Elbow & Wrist & Hip & Knee  & Ankle & Total&\textbf{Refine.} \\
      \hline
      \hline
      
      CMU-Pose~\cite{cao2017realtime} & 91.2  & 87.6  & 77.7  & 66.8  & 75.4  & 68.9 & 61.7 & 75.6 &yes\\
      AE-Pose~\cite{newell2017associative} &92.1  & 89.3  & 78.9  & 69.8  & 76.2  & 71.6 & 64.7 & 77.5 &yes\\
      RefinePose~\cite{fieraru2018learning}& 91.8  & 89.5  & 80.4  & 69.6  & 77.3  & 71.7 & 65.5 & 78.0&yes\\
      PPN~\cite{nie2018pose}&92.2 &89.7&82.1&74.4&78.6&76.4&69.3&80.4&yes\\
      SPM~\cite{nie2019single}&89.7 &87.4 &80.4 &72.4 &76.7 &74.9 &68.3&78.5&yes\\
      \hline
      \hline
      DeepCut~\cite{pishchulin2016deepcut} & 73.4 &71.8 &57.9 &39.9 &56.7 &44.0 &32.0 &54.1&no\\
      DeeperCut~\cite{insafutdinov2016deepercut} &89.4 &84.5 &70.4 &59.3 &68.9 &62.7 &54.6 &70.0&no\\
      Levinkov~\emph{et~al.}~\cite{levinkov2017joint}& 89.8  & 85.2  & 71.8  & 59.6  & 71.1  & 63.0 & 53.5 & 70.6 &no\\
      ArtTrack~\cite{insafutdinov2017arttrack} &88.8  & 87.0  & 75.9  & 64.9  & 74.2  & 68.8 & 60.5 & 74.3&no\\
      %CMU-Pose~\cite{cao2017realtime} & 91.2  & 87.6  & 77.7  & 66.8  & 75.4  & 68.9 & 61.7 & 75.6 &yes& 0.6\\
      RMPE~\cite{fang2017rmpe} & 88.4  & 86.5  & 78.6  & 70.4  & 74.4  & 73.0 & \textbf{65.8} & 76.7 &no\\

      Sekii~\cite{sekii2018pose}&\textbf{93.9} &\textbf{90.2} &\textbf{79.0} &\textbf{68.7}&74.8&68.7&60.5 &76.6&no\\

      \hline
      \textbf{DS-hourglass(mix)}& 91.6  & 88.3  & 78.0  & 68.3  & \textbf{77.9}  & \textbf{72.5} & 65.1 & \textbf{77.4} &no\\

          \hline
  \end{tabular}
  
  \label{tab:table3}
\end{table}

\begin{table}[!t]
 \caption{Results on a set of 288 images from MPII Multi-Person test set.}
  \centering
  \begin{tabular}{c||ccccccc||cc}
  \hline
  Methods &Head & Shoulder & Elbow & Wrist & Hip & Knee  & Ankle & Total&\textbf{Refine.}\\
    \hline
  \hline
  CMU-Pose~\cite{cao2017realtime}& 92.9  & 91.3  & 82.3  & 72.6  & 76.0  & 70.9 & 66.8 & 79.0&yes\\
  AE-Pose~\cite{newell2017associative}&91.5	&87.2&	75.9&	65.4&	72.2&	67.0&	62.1&	74.5&yes\\
  RefinePose~\cite{fieraru2018learning}& 93.8  & 91.6  & 83.9  & 75.2  & 78.0  & 75.6 & 70.7 & 81.3&yes\\
  \hline
  \hline
  DeepCut~\cite{pishchulin2016deepcut}& 73.1  & 71.7  & 58.0  & 39.9  & 56.1  & 43.5 & 31.9 & 53.5 &no\\
  Iqbal and Gall~\cite{iqbal2016multi}& 70.0  & 65.2  & 56.4  & 46.1  & 52.7  & 47.9 & 44.5 & 54.7 &no\\
  
  DeeperCut~\cite{insafutdinov2016deepercut}& 92.1  & 88.5  & 76.4  & 67.8  & 73.6  & 68.7 & 62.3 & 75.6&no\\
  ArtTrack~\cite{insafutdinov2017arttrack}& 92.2  & 91.3  & 80.8  & 71.4  & 79.1  & 72.6 & 67.8 & 79.3&no\\
  
  RMPE~\cite{fang2017rmpe}& 89.4  & 88.5  & 81.0  & 75.4  & 73.7  & 75.4 & 66.5 & 78.6&no\\
  Sekii~\cite{sekii2018pose}&\textbf{95.2} &\textbf{92.2}& \textbf{83.2}&73.8&74.8&71.3&63.4&79.1&no\\
  
    \hline
\textbf{DS-Hourglass(mix)}& 93.2  & 91.1  & 81.5  & \textbf{74.7}  & \textbf{81.0}  & \textbf{75.5} & \textbf{70.2} & \textbf{81.0} &no\\

      \hline
  \end{tabular}

  \label{tab:table4}
\end{table}

%% file: section/conc.tex
\section{Conclusions}
In this paper, we develop a light-weight Hourglass network by applying depthwise separable convolution and mixed depthwise convolution. The new network can be directly applied to single-person pose estimation. Based on this backbone network, we further proposed an efficient multi-person pose estimation method. Both our single-person and multi-person pose estimation methods can achieve competitive accuracies on public datasets with low computational costs.
% \section{Acknowledgements}
% This work was supported in part by the Sichuan Science and Technology Program, China, under grants No.2020YFS0057, and the Fundamental Research Funds for the Central Universities under Project ZYGX2019Z015.